\pdfoutput=1

\documentclass[11pt]{article}

\usepackage{emnlp2021}

\usepackage{times}
\usepackage{latexsym}

\usepackage[T1]{fontenc}

\usepackage[utf8]{inputenc}

\usepackage{microtype}
\usepackage{booktabs}
\usepackage{amsmath}

\usepackage{graphicx}
\usepackage{soul}
\usepackage{url}
\usepackage{arydshln}
\usepackage{algorithm}
\usepackage{algorithmic}
\usepackage{multirow}
\title{Span Fine-tuning for Pre-trained Language Models}

 \author{Rongzhou Bao\textsuperscript{1,2,3}, Zhuosheng Zhang\textsuperscript{1,2,3,}, Hai Zhao\textsuperscript{1,2,3}\\
\textsuperscript{1} Department of Computer Science and Engineering, Shanghai Jiao Tong University\\
\textsuperscript{2} Key Laboratory of Shanghai Education Commission for Intelligent Interaction\\
and Cognitive Engineering, Shanghai Jiao Tong University, Shanghai, China\\
\textsuperscript{3}MoE Key Lab of Artificial Intelligence, Shanghai Jiao Tong University, Shanghai, China\\
\texttt{1154904117@qq.com}\\
\texttt{zhangzs@sjtu.edu.cn,zhaohai@cs.sjtu.edu.cn}\\
}

\begin{document}
\maketitle

\renewcommand{\thefootnote}{\fnsymbol{footnote}}
\footnotetext[1]{ Corresponding author. This paper was partially sup- ported by National Key Research and Development Pro- gram of China (No. 2017YFB0304100), Key Projects of National Natural Science Foundation of China (U1836222 and 61733011). This work was supported by Huawei Noah's Ask Lab} 

\begin{abstract}
Pre-trained language models (PrLM) have to carefully manage input units when training on a very large text with a vocabulary consisting of millions of words. Previous works have shown that incorporating span-level information over consecutive words in pre-training could further improve the performance of PrLMs. However, given that span-level clues are introduced and fixed in pre-training, previous methods are time-consuming and lack of flexibility. To alleviate the inconvenience, this paper presents a novel span fine-tuning method for PrLMs, which facilitates the span setting to be adaptively determined by specific downstream tasks during the fine-tuning phase. In detail, any sentences processed by the PrLM will be segmented into multiple spans according to a pre-sampled dictionary. Then the segmentation information will be sent through a hierarchical CNN module together with the representation outputs of the PrLM and ultimately generate a span-enhanced representation. Experiments on GLUE benchmark show that the proposed span fine-tuning method significantly enhances the PrLM, and at the same time, offer more flexibility in an efficient way. The code is available at https://github.com/BAORONGZHOU/span-fine-tuning.

\end{abstract}

\section{Introduction}

Pre-trained language models (PrLM), including ELECTRA \cite{clark2020electra}, RoBERTa\cite{liu2019roberta}, and BERT \cite{devlin:bert}, have demonstrated strong performance in downstream tasks \cite{wang2018glue}. Leveraging a self-supervised training on large text corpora, these models are able to provide contextualized representations in a more efficient way. For instance, BERT uses Masked Language Modeling and Nest Sentence Prediction as pre-training objects and is trained on a corpus of 3.3 billion words. 

In order to be adaptive for a wider range of applications, PrLMs usually generate sub-token-level representations (words or subwords) as basic linguistic units. For downstream tasks such as natural language understanding (NLU), span-level representations, e.g. phrases and name entities, are also important. Previous works manifest that by changing pre-training objectives, PrLMs' ability to capture span-level information can be strengthened to some extent. For example, base on BERT, SpanBERT \cite{joshi2019spanbert} focuses on masking and predicting text spans, instead of sub-token-level information for pre-training. Entity-level masking is used as a pre-training strategy by ERNIE models \cite{sun2019ernie,zhang2019ernie}. The upper mentioned methods prove that the introduction of span-level information in pre-training to be effective for different NLU tasks.

However, the requirements for span-level information of various NLU tasks differs a lot from case to case. The methods of introducing span-level information in pre-training phase, proposed by previous works, do not fit into the requirements and cannot improve the performance for all NLU tasks. For instance, ERNIE models \cite{sun2019ernie} perform remarkably well in Relation Classification, while underperforms BERT in language inference tasks, such as MNLI \cite{nangia2017repeval}. Therefore, it is imperative to develop a strategy to incorporate span-level information into PrLMs in a more flexible and universally adaptive way. This paper proposes a novel approach, Span Fine-tuning (SF), to leverage span-level information in fine-tuning phase and therefore formulate a task-specific strategy. Compared to existing works, our approach requires less time and computing resources, and is more adaptive to various NLU tasks.

In order to maximize the value and contribution of span-level information, in additional to the sub-token-level representation generated by BERT, Span Fine-tuning also applies a computationally motivated segmentation to further improve the overall experience. Although various techniques, such as dependency parsing \cite{zhou2019limit} or semantic role labeling (SRL) \cite{zhang2019semantics}, have been used as auxiliary tools for sentence segmentation, these methods demand extra parsing procedure, which increase complexities in actual practice. Span Fine-tuning first leverages a pre-sampled $n$-gram dictionary to segment input sentences into spans. Then, the sub-token-level representations within the same span are combined to generate a span-level representation. Finally, the span-level representations are merged with sub-token-representations into a sentence-level representation. In this way, the sentence-level representation is able to contain and maximize the utilization of both sub-token-level and span-level information.

Experiments are conducted on the GLUE benchmark \cite{wang2018glue}, which includes many NLU tasks, such as text classification, semantic similarity, and natural language inference. Empirical results demonstrate that Span Fine-tuning is able to further improve the performance of different PrLMs, including BERT \cite{devlin:bert}, RoBERTa \cite{liu2019roberta} and SpanBERT \cite{joshi2019spanbert}. The result of the experiments with SpanBERT indicates that Span Fine-tuning leverages span-level information differently compared to PrLMs pre-trained with span-level information, which shows the distinguishness of our approach. It is also verified in ablation studies and analysis that Span Fine-tuning is essential for further performance improvement for PrLMs.

\section{Related Work}
\subsection{Pre-trained language models}
Learning reliable and broadly applicable word representations has been an ongoing heated focus for natural language processing community. Language modeling objectives are proved to be effective for distributed representation generation \cite{mnih2009scalable}. By generating deep contextualized word representations, ELMo \cite{Peters2018ELMO} advances state of the art for several NLU tasks. Leveraging Transformer \cite{vaswani2017attention}, BERT \cite{devlin:bert} further advances the field of transfer learning. Recent PrLMs are established based on the various extensions of BERT, including using GAN-style architecture \cite{clark2020electra}, applying a parameter sharing strategy \cite{lan2019albert}, and increasing the efficiency of parameters \cite{liu2019roberta}. 
\begin{figure*}[t]
	\centering
	\includegraphics[width=0.95\textwidth]{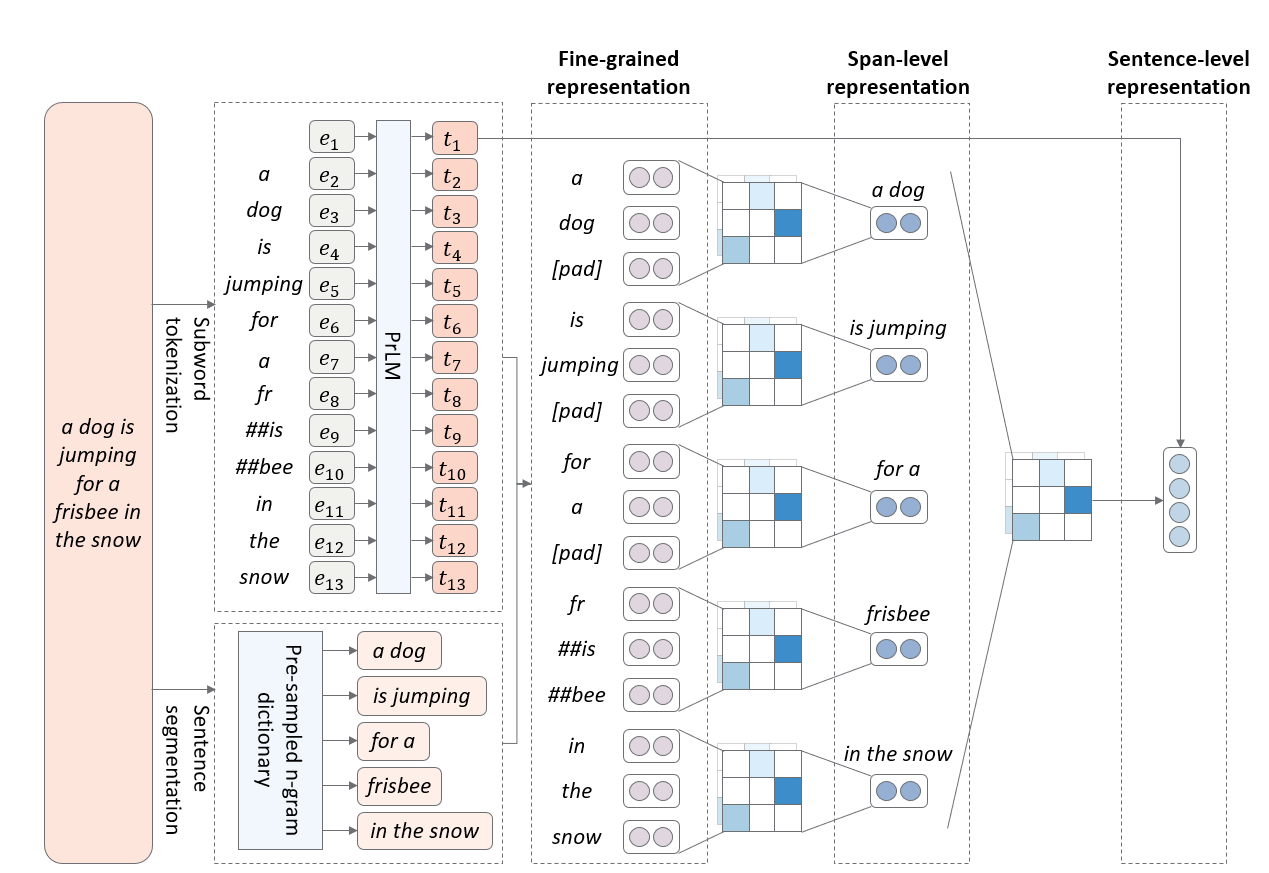}
	\caption{\label{fig:framework} Overview of the  framework of our proposed method}
\end{figure*}
\subsection{Span-level pre-training methods}
Previous works manifest that the introduction of span-level information in pre-training phase can improve PrLMs' performance. In the first place, BERT leverages the prediction of single masked tokens as one of the pre-training objectives. Due to the use of WordPiece embeddings \cite{wu2016google}, BERT is able to segment sentences into sub-word level tokens, so that the masked tokens are at sub-token-level, e.g. "\#\#ing". \cite{devlin:bert} shows that masking the whole word, rather than only single tokens, can further enhance the performance of BERT. Later, it is proved by \cite{sun2019ernie,zhang2019ernie} that the masking of entities is also helpful for PrLMs. By randomly masking adjoining spans in pre-training, SpanBERT \cite{joshi2019spanbert} can generate better representation for given texts. AMBERT \cite{zhang2020ambert} achieves better performance than its precursors in NLU tasks by incorporating both sub-token-level and span-level tokenization in pre-training. The upper mentioned studies all focus on introducing span-level information in pre-training. To the best of our knowledge, the introduction of span-level information in fine-tuning is still a white space to explore, which makes our approach a valuable attempt.

\subsection{Integration of fine-grained representation}

Different formats of downstream tasks require sentence-level representations, such as natural language inference \citep{Bowman2015A,nangia2017repeval}, semantic textual similarity \citep{cer2017semeval} and sentiment classification \citep{socher2013recursive}. Besides directly pre-training the representation of coarser granularity \citep{le2014distributed,logeswaran2018efficient}, a lot of methods have been explored to obtain a task-specific sentence-level representation by integrating  fine-grained token-level representations\citep{conneau2017supervised}. \citet{kim2014convolutional} shows that by applying a convolutional neural network (CNN) on top of pre-trained word vectors, we can get a sentence-level representation that is well adapted to classification tasks.  \citet{lin2017structured} leverage a self-attentive module over hidden states of a BiLSTM to generate sentence-level representations. \citet{zhang2019semantics} use a CNN layer to extract word-level representations form sub-word representations and combine them with word-level semantic role representations. Inspired by these methods, after a series of preliminary attempts, we choose a hierarchical CNN architecture to recombine fine-grained representations to coarse-grained ones.

\section{Methodology }

Figure \ref{fig:framework} demonstrates the overall framework of Span Fine-tuning, which is essentially uses BERT as a foundation and incorporates segmentation as an auxiliary tool. The figure does not exhaustively depict the details of BERT, given the model is relatively popular and ubiquitous. Further information on BERT is available in \cite{devlin:bert}. In Span Fine-tuning, an input sentence is divided into sub-word-level tokens and then sent to BERT to generate sub-token-level representations. At the same time, the input is segmented into spans based on $n$-gram statistics. By combining the segmentation information with sub-token-level representations generated by BERT, we divided the representation into several spans. Then, the spans are sent through a hierarchical CNN module to obtain a span-level information enhanced representation. Finally, the sub-token-level representation of \texttt{[CLS]} token generated by BERT and the span-level information enhanced representation are concatenated and form a final representation, which maximized the value of both sub-token-level and span-level information for NLU tasks.
\begin{figure*}[t]
	\centering
	\includegraphics[width=0.95\textwidth]{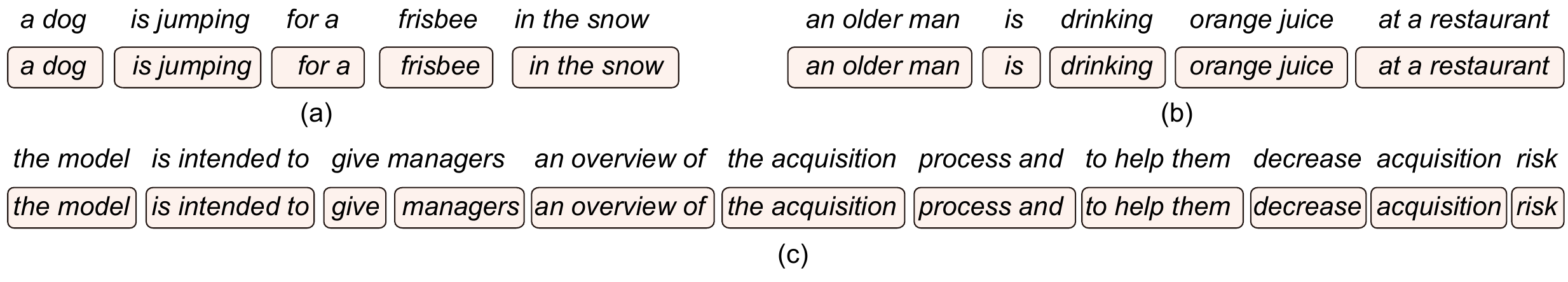}
	\caption{\label{fig:segmentation} Segmentation Examples }
\end{figure*}
\subsection{Sentence Segmentation}
Semantic role labeling (SRL) \cite{zhang2019semantics} and dependency parsing \cite{zhou2019limit} have been used as auxiliary tools for segmentation by previous works. Nonetheless, these techniques demand additional parsing procedures, and therefore increase complexities for real application. In order to obtain a simpler and more convenient segmentation, base on frequency, we select meaningful $n$-grams appeared in wikitext-103 dataset\footnote{PMI method has also been tried to adjust our dictionary, but the result is not competitive.} to form a pre-sampled dictionary.

We use the dictionary to match $n$-grams from the head of each input sentence. $n$-grams with greater lengths are prioritized, while unmatched tokens remain the same. In this way, we are able to obtain a specific segmentation of the input sentence. Figure \ref{fig:segmentation} demonstrates some examples of sentence segmentation from the GLUE dataset.

\subsection{Sentence Encoder Architecture}
An input sentence $X=\{x_1, \dots, x_n\}$ is given with a length $n$. The sentence is firstly divided into sub-word tokens (with a special token \texttt{[CLS]} at the beginning) and converted to sub-token-level representations $E=\{e_1, \dots, e_m\}$ (usually $m$ is larger than $n$) according to embeddings proposed by \cite{wu2016google}. Then, the transformer encoder (BERT) captures the contextual information for each token by self-attention and generates a sequence of sub-token-level contextual embeddings $T=\{t_1, \dots, t_m\}$, in which $t_1$ is the contextual representation of special token \texttt{[CLS]}. Based on the segmentation generated by the $n$-gram statistics, the sub-token-level contextual representations are combined into several spans $\{C_1, \dots, C_r\}$, with $r$ as a hyperparameter indicating the max number of spans for all processed sentences. Each $C_i$ contains several contextual sub-token-level representations extracted from $T$ dedoted as $\{t^i_1, t^i_2, ..., t^i_l\}$. $l$ is another hyperparameter representing the max number of tokens for all the spans. A CNN-Maxpooling module is applied to each $C_i$ to get a span-level representation $c_i$:
\begin{equation}
\begin{split}
c{^i_j} &= ReLU(W_1\left[t^i_j, t^i_{j+1}, \dots,t^i_{j+k-1}\right] + b_1),\\
c_i &= MaxPooling({c^i_1, \dots,c^i_r}),
\end{split}
\end{equation}
where $W_1$ and $b_1$ are trainable parameters and $k$ is the kernel size. Based on the span-level representations $\{c_1, \dots, c_r\}$, another CNN-Maxpooling module is applied to obtain a sentence-level representation $s$ with enhanced span-level information:
\begin{equation}
\begin{split}
s{'_i} &= ReLU(W_2\left[c_i, c_{i+1}, \dots,c_{i+k-1}\right] + b_2),\\
s &= MaxPooling({s{'_1}, \dots,s{'_r}}),
\end{split}
\end{equation}
Finally, we concatenate $s$ with the contextual sub-token-level representation $t_1$ of special token \texttt{[CLS]} provided by BERT, and generate a sentence-level representation $s^*$ that maximizes the value of both sub-token-level and span-level information for NLU tasks:
$s^* = s \diamond t_1$.

\begin{table*}
	\centering
	\setlength{\tabcolsep}{2pt}
	{
		\begin{tabular}{lcccccccccc}
			\hline
			
			\hline
			
			Method& CoLA & SST-2 & MNLI & QNLI & RTE  & MRPC  & QQP & STS-B & Avg & Gain\\ 
			&  (mc) & (acc)	& m/mm(acc) & (acc) & (acc)  & (F1) & (F1) & (pc) & - & -\\
			\hline
			\multicolumn{10}{c}{\emph{In literature}} \\
			BERT$_\text{BASE}$  & 52.1 & 93.5  & 84.6/83.4 &  - & 66.4 & 88.9 & 71.2  & 87.1 & 78.3\\
			BERT$_\text{LARGE}$ &60.5 & \textbf{94.9} &  86.7/85.9 & 92.7 & 70.1  & 89.3 & 72.1 & 87.6 & 80.5\\
			\hdashline
			BERT-1seq\footnotemark[3]& 63.5 & 94.8 & 88.0/87.4 & 93.0 & 72.1 & 91.2 & 72.1 & 89.0 &  83.5 &\multirow{2}{*}{1.0}\\
		    SpanBERT & \textbf{64.3} & 94.8 & \textbf{88.1}/\textbf{87.7} & \textbf{94.3} & 79.0 & 90.9 & 71.9 & \textbf{89.9} &  \textbf{84.5}\\
			\hline
			\multicolumn{10}{c}{\emph{Our implementation}} \\
			BERT$_\text{BASE}$ & 51.4 & 92.1 & 84.4/83.5 & 90.3 & 67.1 & 88.3 & 71.3 & 85.1 &  79.3 & \multirow{2}{*}{\textbf{1.1}}\\
			BERT$_\text{BASE}$ + SF & 55.1 & 93.6 & 84.8/84.3 & 90.6 & 69.6 & 88.7 & 71.9 & 86.5 &  80.4\\
			\hdashline
			BERT$_\text{WWM}$   & 61.1 &93.6 &87.1/86.5 & 93.9 & 77.3  & 90.0 & 71.9 & 88.1 &  83.3 & \multirow{2}{*}{\textbf{1.1}}\\
			BERT$_\text{WWM}$ + SF   & 62.9 &94.1 &87.6/87.0 & \textbf{94.3} & \textbf{81.4}  & \textbf{91.1} & \textbf{72.4} & 89.1 &  84.4\\
		
			\hline
			
			\hline
		\end{tabular}
	}
	
	\caption{\label{tab:glue} Test sets performance on GLUE benchmark. All the results are obtained from \cite{liu2019multi}, \cite{radford2018improving}. For a simple demonstration, problematic WNLI set are excluded, and we do not show the accuracy of the datasets have F1 scores. \emph{mc} and \emph{pc} denote the Matthews correlation and Pearson correlation respectively. 
	}
	
\end{table*}

\subsection{Tasks and Datasets}
  To evaluate Span Fine-tuning, experiments are conducted on nine NLU benchmark datasets, including text classification, natural language inference, semantic similarity. Eight of which are available from the GLUE benchmark \cite{wang2018glue}. And the rest one is SNLI \cite{Bowman2015A}, a widely accepted natural language inference dataset. 

\subsection{Pre-trained Language Model}
We leverage the PyTorch implementation of BERT \cite{devlin:bert}, RoBERTa \cite{liu2019roberta} and SpanBERT \cite{joshi2019spanbert} based on HuggingFace’s codebase\footnote{\url{https://github.com/huggingface}} \cite{Wolf2019HuggingFacesTS}  as our PrLMs and baselines. 

\section{Experiments}
\subsection{Set Up}
We select all the $n$-grams with $n \le 5$, which occurs more than ten times in the wikitext-103 dataset, to form a dictionary. The pre-sampled dictionary, containing more than 400 thousand $n$-grams, is used to segment input sentences. During segmentation, two hyperparameters are involved: $r$ representing the largest number of spans in a sentence, and $l$ indicating the largest number of tokens included in a span. In order to maintain different dimensions of features for each input sentence, padding and tail are employed. We set $r$ equals to 16, and based on NLU tasks, choose $l$ in \{64,128\} .  

The fine-tuning procedure is as the same as BERT's. Adam is used as the optimizer. The initial learning rate is in \{1e-5,2e-5, 3e-5\}, the warm-up rate is 0.1, and the L2 weight decay is 0.01. The batch size is set in \{16, 32, 48\}. The maximum number of epochs is set in \{2,3,4,5\} based on NLU tasks. Input sentences are divided into subtokens and converted to WordPiece embeddings, with a maximum length in \{128, 256\}. The output size of the CNN layer is the same as the hidden size of PrLM, and the kernel size is set to 3.

\subsection{Results with BERT as PrLM}
Two released BERT \cite{devlin:bert}, BERT Large Whole Word Masking and BERT Base, are first used as pre-trained encoder and baselines for Span Fine-tuning. Compared with BERT Large, BERT Large Whole Word Masking reach a better performance, since it uses whole-word masking in pre-training phase. Therefore, we select BERT Large Whole Word Masking as a stronger baseline. The results indicate that Span Fine-tuning can maximize the contribution of span-level information, even when compared to a stronger baseline.

Table \ref{tab:glue} exhibits the results on the GLUE datasets, showing that Span Fine-tuning can significantly improve the performance of PrLMs. Since our approach leverages BERT as a foundation, and undergoes the the same evaluation procedure, it is evident that the performance gain is fully contributed by the newly introduced Span Fine-tuning.

In order to test the statistical significance of the results, we follow the procedure of  \cite{zhang2020retrospective}. We use the McNemars test, this test is designed for paired nominal observations, and it is appropriate for binary classification tasks.The p-value is defined as the probability of obtaining a result equal to or more extreme than what was observed under the null hypothesis. The smaller the p-value, the higher the significance. A commonly used level of reliability of the result is 95\%, written as p = 0.05.  As shown in table \ref{tab:pvalue}, compared with the baseline, for all the binary classification tasks of GLUE benchmark, our method pass the significance test.

\begin{table}[htb!]
	\centering\small
	\setlength{\tabcolsep}{3pt}
	{
		\begin{tabular}{lcccccc}
			\toprule
			& CoLA & SST-2 & QNLI& RTE& MRPC&QQP\\
			\midrule
			p-value & 0.005 & 0.012 & 0.023 & 0.009& 0.008&0.031\\
			\bottomrule
		\end{tabular}
	}
	
	\caption{\label{tab:pvalue} Results of McNemars tests for binary classification tasks of GLUE benchmark, tests are conducted based on the results of best run of BERT$_\text{BASE}$ and BERT$_\text{BASE}$ + SF.
	}
	
\end{table}

\footnotetext[3]{The baseline of SpanBert, a BERT pre-trained without next sentence prediction object.}

Span Fine-tuning can reach the same performance improvement as previous methods. As illustrated in Table \ref{tab:glue}, on average, SpanBERT can improve the result by one percentage point over the baseline (BERT-1seq), while Span Fine-tuning is able to achieve an improvement of 1.1 percentage points over our baseline. However, as showed in Table \ref{tab:compare}, Span Fine-tuning requires considerably less time and computing resources compared to the large-scale pre-training for span-level information incorporation. When the Span Fine-tuning is adopted, the extra parameters are only 3 percent of the total parameters of the adopted PrLMs for every downstream task, and introduce little extra overhead.

\begin{table}[htb!]
\centering
\begin{tabular}{lcc}
\toprule
Method  & Time & Resource\\
\midrule
Pre-train  & 32 days& 32 Volta V100  \\
Span Fine-tune & 12 hours max & 2 Titan RTX\\

\bottomrule
\end{tabular}
\caption{\label{tab:ablation} The comparison between incorporation of span-level information in pre-training and Span Fine-tuning .}
\label{tab:compare}
\end{table}

\begin{table*}
	\centering
	\setlength{\tabcolsep}{5pt}
	{
		\begin{tabular}{lccccccccc}
			\toprule
			
			Method& CoLA & SST-2 & MNLI & QNLI & RTE  & MRPC  & QQP & STS-B & Avg.\\ 
			&  (mc) & (acc)	& m/mm(acc) & (acc) & (acc)  & (F1) & (acc) & (pc) & -\\
			\midrule
			SpanBERT$_\text{LARGE}$ & 64.3 & 94.8 & 88.1/87.7 & 94.3 & 79.0 & 90.9 & 89.5 & 89.9 & 86.5\\
			SpanBERT$_\text{LARGE}$ + SF & 65.9 & 95.1 & 88.4/88.1 & 94.3 & 83.3 & 92.1 & 90.9 & 90.1 & 87.6\\
			RoBERTa$_\text{LARGE}$ & 68.0 & \textbf{96.4} & 90.2/\textbf{90.2} & \textbf{94.7} & 86.6 & 90.9 & \textbf{92.2} & \textbf{92.4} & 89.0\\
			RoBERTa$_\text{LARGE}$ + SF  & \textbf{68.9} &96.1 &\textbf{90.3}/\textbf{90.2} & 94.3 & \textbf{90.6}  & \textbf{92.8} & \textbf{92.2} & \textbf{92.4} &  \textbf{89.8}\\
			\bottomrule
		\end{tabular}
	}
	
	\caption{\label{tab:strong} Results on test sets of GLUE benchmark with stronger baseline, we average results from three different random seeds.
	}
	
\end{table*}

Besides, Span Fine-tuning is more flexible and adaptive compared to previous methods. Table \ref{tab:glue} shows that Span Fine-tuning is able to achieve stronger results on all NLU tasks compared to the baseline, whereas the results of SpanBERT in certain tasks, such as Quora Question Pairs and Microsoft Research Paraphrase Corpus, are worse than its baseline. Since for spanBERT, the utilization of span-level information is fixed for every downstream task. Whereas in our method, an extra module designed to incorporate span-level information is trained during the fine-tuning, which can be more dynamically adapted to different downstream tasks.

\begin{table}[htb!]
\centering
\begin{tabular}{lrr}
\toprule
Method  & Dev & Test \\
\midrule
BERT$_\text{WWM}$   & 92.0& 91.4      \\
BERT$_\text{WWM}$ + SF&92.3& 91.7\\
SemBERT$_\text{WWM}$&92.2& 91.9\\
\bottomrule
\end{tabular}
\caption{\label{tab:snli}  Accuracy on dev and test sets of SNLI. SemBERT$_\text{WWM}$ \cite{zhang2019semantics} is the published SoTA on SNLI.}
\label{tab:booktabs}
\end{table}

Table \ref{tab:snli} indicates that Span Fine-tuning also enhances the result of PrMLs on the SNLI benchmark. 
The improvement achieved by Span Fine-tuning is similar to published state-of-the-art accomplished by SemBERT. However, compared to SemBERT, Span Fine-tuning saves a lot more time and computing resources. Span Fine-tuning merely leverages a pre-sampled dictionary to facilitate segmentation, whereas SemBERT leverages a pre-trained semantic role labeller, which brings extra complexities to the whole segmentation process. 

Furthermore, Span Fine-tuning is different from SemBERT in terms of motivation, method and contribution factors. The motivation of SemBERT is to enhance PrLMs by incorporating explicit contextual semantics, whereas the motivation of our work is to let PrLMs leverage span-level information in fine-tuning. When it comes to method,  SemBERT concatenate the original representations given by BERT with representations of semantic role labels, in comparison, our work directly leverages a segmentation given by a pre-sampled dictionary to generate span-enhanced representation and requires no pre-trained semantic role labeler.  The gain of SemBERT comes from semantic role labels while the gain of our work comes from the specific segmentation, which is very different. 

It's worth noticing that semantic role labeler can also generate segmentation. However, semantic role labeler will generate multiple segmentation for sentence which has various predicate-argument structures. Furthermore, such segmentation is sometimes coarse-grained (with span more than ten words), which is unpractical for our work.

\subsection{Results with Stronger PrLMs}

In addition to BERT, we also apply Span Fine-tuning to stronger PrLMs, such as RoBERTa \cite{liu2019roberta} and SpanBERT \cite{joshi2019spanbert}, which optimize BERT by enhancing pre-training procedure and predicting text spans rather than single tokens respectively.

Table \ref{tab:strong} shows that Span Fine-tuning can strengthen both RoBERTa and SpanBERT. RoBERTa is a already very strong baseline, we remarkably improve the performance of RoBERTa on RTE by four percentage points. SpanBERT already incorporated span-level information during the pre-training, but the results still support that Span Fine-tuning utilizes the span-level formation and improves the performance of PrLMs in a different dimension.

\section{ Analysis}
\subsection{Ablation Study}

In order to determine the key factors in Span Fine-tuning, a series of studies are conducted on the dev sets of eight NLU tasks. BERT$_\text{BASE}$ is chosen as the PrLM for the ablation studies. As shown in Table \ref{tab:ablation}, three sets of ablation studies are performed. For experiment  BERT$_{BASE}$ + CNN, only a hierarchical CNN structure is applied in  to evaluate whether the improvement comes from the extra parameters. To illustrate, we firstly apply two layers of CNN over the token-level representations given by BERT. Then, a max pooling operation is applied to get the sentence-level representation. Finally, the sentence-level representation and the 'CLS' representation of BERT is concatenated and sent to the classifier. In this way, the parameters of  BERT$_{BASE}$ + CNN are the same as in our method.  For experiment BERT$_{BASE}$ + CNN + Random SF, random sentence segmentation is applied to the experiment to test if the proposed segmentation method of Span Fine-tuning really functions in span-level information incorporation.  For experiment BERT$_{BASE}$ + CNN + NLTK SF, we conduct the experiments using a pre-trained chunker from Natural Language Toolkit to see whether the proposed segmentation method of Span Fine-tuning can achieve further improvements.

\begin{table}[h!]
\centering
\begin{tabular}{lc}
\toprule
method  & Avg Score \\
\midrule
BERT$_\text{BASE}$   & 82.6      \\
BERT$_\text{BASE}$ + CNN&82.5\\
BERT$_\text{BASE}$ + Random SF\footnotemark[4] &    83.0\\
BERT$_\text{BASE}$ + NLTK SF\footnotemark[5] &   83.7\\
BERT$_\text{BASE}$ + SF &   \textbf{84.2}\\
\bottomrule
\end{tabular}
\caption{\label{tab:ablation}  Ablation studied on dev sets of GLUE benchmark, we average results from three different random seeds.}
\end{table}
\footnotetext[4]{Random SF represents Span Fine-tuning with randomly segmented sentences.}
\footnotetext[5]{NLTK SF represents Span Fine-tuning with segmentation generated by an NLTK pre-trained chunker.}

The results of the experiment BERT$_{BASE}$ + CNN suggest that the improvement is unlikely to come from the extra parameters, since it reduce the overall performance by 0.1 percent. The experiment BERT$_{BASE}$ + Random SF and BERT$_{BASE}$ + NLTK SF indicate that the segmentation generated by a pre-train chunker or even random segmentation can also achieve enhancement under the Span Fine-tuning structure. However, a pre-trained chunker demands additional part-of-speech parsing process, while our segmentation method only relies on a pre-sampled dictionary and saves a lot more time, and at the same time, achieves greater improvement. Our Span Fine-tuning is able to remarkably enhance the result on all NLU tasks, raising average score by 1.6 percentage points. Overall, the result of experiments indicate that the performance improvement is primarily a result of our unique segmentation method.

\subsection{Encoder Architecture}

\cite{conneau2017supervised} mentions that the influence of sentence encoder architectures on PrLM performance varies a lot from case to case. \cite{toshniwal-etal-2020-cross} also suggests that different span representations can affect NLPs tasks greatly.

\begin{table}[htb!]
\centering
\begin{tabular}{lrr}
\toprule
Method  & Dev & Test \\
\midrule
CNN-Max   & 90.9& 90.9      \\
CNN-CNN&\textbf{91.3}& \textbf{91.1}\\
 Attention\footnotemark[6]-Max &90.7& 90.5\\
 Attention-Attention&90.8& 90.8\\
\bottomrule
\end{tabular}
\caption{\label{tab:structure}  Accuracy on dev and test sets of SNLI. SemBERT$_\text{WWM}$ \cite{zhang2019semantics} is the published SOTA on SNLI.}
\label{tab:booktabs}
\end{table}
\footnotetext[6]{Attention indicate the Self-attentive module \cite{lin2017structured}.}

To evaluate the effectiveness of our encoder architecture, we replace the component of the encoding layer and the overall structure respectively. For the component of the encoding layer, CNN \cite{kim2014convolutional} and the Self-attentive module \cite{lin2017structured} are compared. For the overall structure, two structures are considered: a single layer structure with the max-pooling operation and a hierarchical structure.

By matching every component of the encoding layer with the overall structure, four different encoder architectures are generated: CNN-Maxpooling, CNN-CNN, Attention-Maxpooling, Attention-Attention.
Experiments are conducted on SNLI dev and test sets. Table \ref{tab:structure} suggests that the hierarchical CNN (CNN-CNN) is most suitable encoder architecture for us.

\subsection{Size of $n$-gram Dictionary}
\begin{figure}[htb!]
	\centering
	\includegraphics[width=0.48\textwidth]{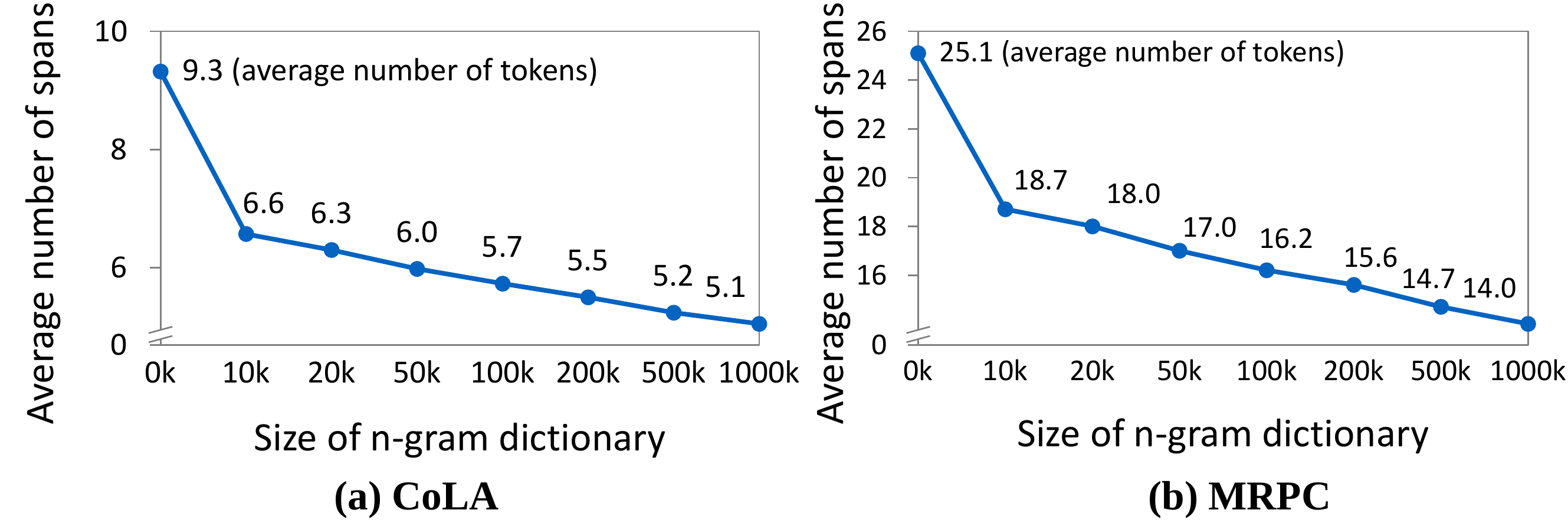}
	\caption{\label{fig:span_number} Influence of dictionary size on the average number of spans in the sentences}
\end{figure}

Since our segmentation method is based on a pre-sampled dictionary, the size of dictionaries will have a large impact on segmentation results. Figure \ref{fig:span_number} depicts how the average number of spans in the sentences changed along with dictionary size in CoLA and MRPC datasets. At the origin, where no segmentation is applied, every token is considered as a span. The number of spans drops significantly, as the dictionary size grows and more $n$-grams are matched and grouped together.

\begin{figure}[htb!]
	\centering
	\includegraphics[width=0.48\textwidth]{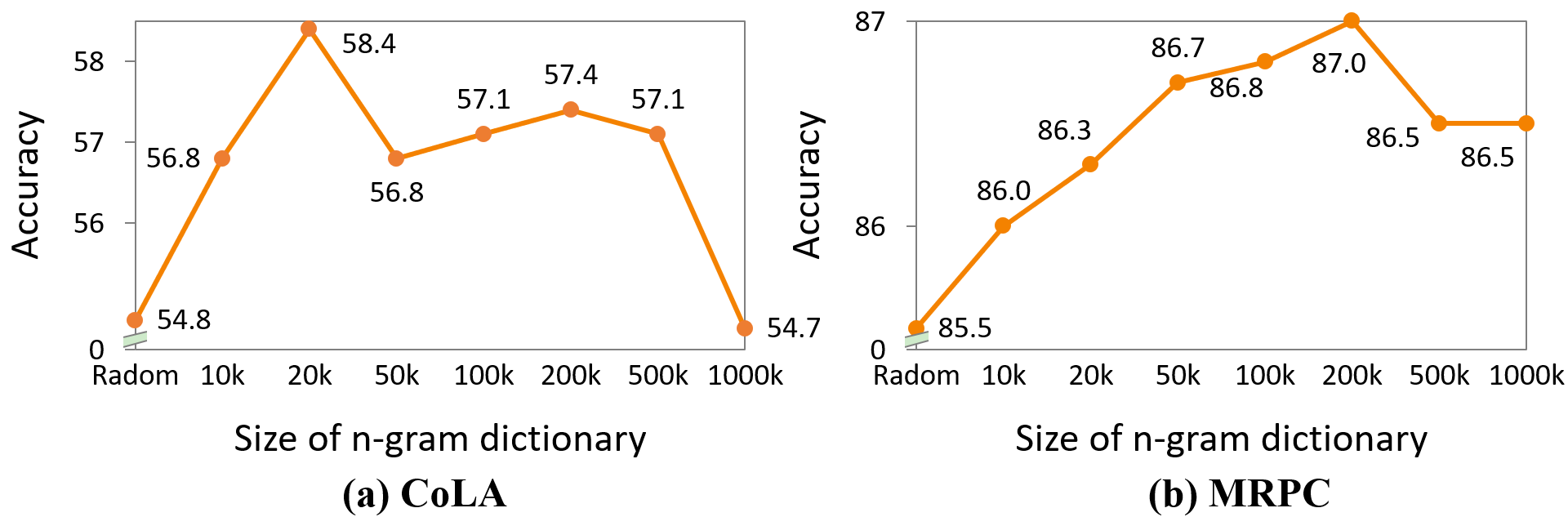}
	\caption{\label{fig:case_study} The influence of the size of $n$-gram dictionary}
\end{figure}

To evaluate the influence of dictionary size on PrLM performance, experiments on the dev sets of two NLU tasks are implemented: CoLA and MRPC. To concentrate on the impact of segmentation and reduce the impacts from sub-token-level representations provided by PrLM, the concatenation process is not applied to this experiment. Rather, the span-level information enhanced representations are directly sent to a dense layer to generate prediction. As demonstrated in figure \ref{fig:case_study}, the incorporation of pre-sampled $n$-gram dictionary generates a stronger performance compared to random segmentation. Moreover, dictionaries of medium sizes (20$k$ to 200$k$) commonly result in better performance. Such trend can be explained by intuition, give dictionaries of small sizes are likely to omit meaningful $n$-grams, whereas the ones of large sizes tend to over-combine meaningless $n$-grams.

\subsection{Span Fine-tuning for Token-Level Tasks}
The upper mentioned experiments are conducted on the GLUE benchmark, whose tasks are all at the sentence level. Nevertheless, token-level representations are needed in many other NLU task, such as name-entity recognition (NER). Our approach can be applied to token-level tasks with simple modification of encoder architecture (e.g. removing the pooling layer of CNN module). Table \ref{tab:ner} shows the results of our approach on the CoNLL-2003 Named Entity Recognition (NER) task \cite{tjong-kim-sang-de-meulder-2003-introduction} with BERT as our PrLM.

\begin{table}[htb!]
	\centering\small
	\setlength{\tabcolsep}{1pt}
	{
		\begin{tabular}{lccccc}
			\toprule
			& BERT$_\text{BASE}$ & BERT$_\text{BASE}$+SF& BERT$_\text{LARGE}$ & BERT$_\text{LARGE}$+SF \\
			\midrule
			Dev & 91.7 & 92.1 & 92.3 & 92.5 \\
		    Test & 95.7 & 96.2 & 96.5 & 96.8 \\
			\bottomrule
		\end{tabular}
	}
	
	\caption{\label{tab:ner}  F1 on dev and test sets of named entity recognition from
    CoNLL-2003, we average results from three different random seeds.
	}
	
\end{table}

\section{Conclusion}
This paper proposes Span Fine-tuning that maximize the advantages of flexible span-level information in fine-tuning with sub-token-level representations generated by PrLMs. Leveraging a reasonable segmentation provided by a pre-sampled $n$-gram dictionary, Span Fine-tuning can further enhance the performance of PrLMs on various downstream tasks. Compared with previous span pre-training methods, our Span Fine-tuning remains competitive for the following reasons:

\paragraph{Task-adaptive}
For methods that incorporate span-level information in pre-training, the utilization of span-level information is unlikely easily adjusted for every downstream task as span pre-training has been fixed after tremendous computational cost. In our method, the extra module designed to incorporate span-level information is trained during the fine-tuning, resulting in a more dynamically adaptation to different downstream tasks.
\paragraph{Flexible to PrLMs}
Our approach can be generally applied to various PrLMs including RoBERTa and SpanBERT.
\paragraph{Novelty}
Our approach can further improve the performance of PrLMs pre-trained with span-level information (e.g. SpanBERT). Such result implies that we our method utilizes the span-level information in a different manner comparing with PrLMs pre-trained with span-level information, which makes our method distinguished comparing with previous works.

\bibliography{anthology,custom}
\bibliographystyle{acl_natbib}

\end{document}